# Atomic Fact-Checking Increases Clinician Trust in Large Language Model Recommendations for Oncology Decision Support: A Randomized Controlled Trial

Lisa C. Adams[1], Linus Marx[2], Erik Thiele Orberg[3,4], Keno Bressem[5], Sebastian Ziegelmayer[1], Denise Bernhardt[2], Markus Graf[1], Marcus R. Makowski[1], Stephanie E. Combs[2,6], Florian Matthes[7], Jan C. Peeken[2,6]

1 Department of Diagnostic and Interventional Radiology, TUM University Hospital Rechts der Isar, TUM School of Medicine and Health, Technical University of Munich, Munich, Germany
2 Department of Radiation Oncology, TUM University Hospital Rechts der Isar, TUM School of Medicine and Health, Technical University of Munich, Munich, Germany
3 Department of Internal Medicine III: Hematology and Clinical Oncology, University Hospital Regensburg, University of Regensburg, Regensburg, Germany
4 Bavarian Cancer Research Center (BZKF), Regensburg, Germany
5 Institute for Cardiovascular Radiology and Nuclear Medicine, TUM University Hospital, German Heart Center Munich, TUM School of Medicine and Health, Technical University of Munich, Munich, Germany
6 German Consortium for Translational Cancer Research (DKTK), Partner Site Munich, Munich, Germany
7 Department of Computer Science; TUM School of Computation Information and Technology; Technical University of Munich, Garching, Germany

Corresponding Author: Jan C. Peeken, MD PhD
Email: jan.peeken@tum.de
Address: Department of Radiation Oncology, TUM School of Medicine and Health, TUM University Hospital Rechts der Isar, Munich, Germany

**Abstract**

**Background:** Large language models (LLMs) can synthesize clinical guidelines and generate diagnostic and treatment recommendations, yet clinician trust remains a barrier to adoption. Traditional approaches emphasize natural language explanations of LLM-aided recommendations and source citations, but their effectiveness in high-stakes clinical settings is uncertain.

**Objective:** To determine whether atomic fact-checking (AFC), which decomposes AI recommendations into individually verifiable claims linked to source guideline documents, increases clinician trust compared to explanations, citations, and other traditional transparency interventions.

**Methods:** A randomized controlled trial was conducted with 356 clinicians (160 radiologists, 85 radiation oncologists, 111 medical oncologists). Participants evaluated AI-generated recommendations for 21 oncology cases across seven cancer types (prostate, breast, lung, colorectal, kidney, liver, lymphoma), yielding 7,476 trust ratings. Participants were randomly assigned to one of two arms (with or without natural language explanations) and further sub-randomized to one of five transparency conditions: (1) recommendation only, (2) recommendation with explanation, (3) recommendation with source citation, (4) recommendation with explanation and citation, or (5) recommendation with explanation, citation, and AFC. The primary outcome was trust measured on a validated 5-point Likert scale.

**Results:** AFC produced significantly higher trust than all control conditions. Mean trust scores were 2.59 (95% CI, 2.54-2.64) for recommendation only, 2.84 (2.79-2.90) for explanation, 3.01 (2.97-3.05) for source citation, 3.09 (3.04-3.14) for combined explanation and citation, and 3.80 (3.76-3.84) for AFC. The effect size for AFC versus pooled controls was Cohen's d = 0.94 (95% CI, 0.88-1.00; $P < .001$). The proportion expressing trust (score $\geq$ 4) increased from 26.9% to 66.5% with AFC (absolute increase: 39.5 percentage points; number needed to treat: 2.53). Effects were consistent across specialties (d = 0.80-1.03), cancer types (d = 0.79-1.10), and experience levels (d = 0.62-1.41).

**Conclusions:** AFC substantially increases clinician trust in AI-generated oncology recommendations. Decomposing AI outputs into verifiable claims with linked guideline sources produces larger effects than traditional transparency mechanisms.

**Key Points**

**Question:** Does atomic fact-checking, which decomposes AI treatment recommendations into individually verifiable claims linked to source guideline documents, increase clinician trust compared to traditional explainability approaches?

**Findings:** In this randomized trial of 356 clinicians generating 7,476 trust ratings, atomic fact-checking produced a large effect on trust (Cohen's d = 0.94), increasing the proportion of clinicians expressing trust from 26.9% to 66.5%. Traditional transparency mechanisms showed a dose-response gradient of improvement over baseline (d = 0.25 to 0.50).

**Meaning:** Decomposing AI recommendations into individually verifiable claims linked to source guidelines produces substantially higher clinician trust than traditional explainability approaches in high-stakes clinical decisions.

# Introduction

Large language models (LLMs) have demonstrated capabilities in medical knowledge synthesis, clinical reasoning, and treatment recommendation generation that match or exceed prior AI systems in medicine (1-4). These advances have raised expectations for AI-augmented clinical decision-making, yet the gap between technical capability and clinical deployment remains a central challenge in digital medicine (5).

However, translation into clinical practice has been slower than anticipated (6). A scoping review of randomized controlled trials that evaluate the clinical application of AI found that while 81% reported positive primary endpoints, trust remains a persistent barrier (7).

Clinician trust has emerged as a critical factor in this translation gap. Trust is not merely a psychological construct but a practical prerequisite for clinical utility: clinicians who distrust AI recommendations will not incorporate them into decision-making, regardless of underlying accuracy (8, 9). Conversely, excessive or miscalibrated trust poses patient safety risks. The challenge, therefore, is to develop mechanisms that enable appropriate, calibrated trust—a prerequisite for responsible deployment of AI-based decision support in digital health systems.

Traditional approaches to building trust in AI systems have focused on transparency, typically through two mechanisms: explanation generation and source citation (10, 11). These approaches assume that understanding how an AI system reaches its conclusions enables appropriate trust calibration. Substantial research effort has been devoted to developing explainable AI (XAI) such as natural language reasoning chains (12). Yet empirical evidence for the effectiveness of these transparency mechanisms in clinical settings remains limited and sometimes contradictory. Gombolay and colleagues, in a randomized study of explainable AI in neurology decision support, found that different XAI methods had variable impacts on clinician performance and that increasing perceived explainability paradoxically degraded performance among experienced clinicians (13). A population-level preference study by Ploug et al. found that while explainability was valued, physician responsibility for final decisions was considered more important (14). These findings suggest that explanation-based approaches may be insufficient for building clinical trust in high-stakes settings.

An alternative approach addresses this limitation directly: decomposing AI recommendations into discrete, verifiable claims that clinicians can independently confirm against source evidence (15). Atomic fact-checking (AFC) extracts individual factual claims from AI-generated recommendations and presents each claim with a verification status indicator and a direct link to the corresponding passage in the source clinical guideline. The guideline source is displayed alongside the decomposed claims, enabling clinicians to confirm that each atomic fact accurately reflects guideline content.

The theoretical basis for AFC rests on the distinction between trusting AI reasoning versus trusting one's own verification. Traditional explainability asks clinicians to evaluate AI reasoning, a cognitively demanding task (13, 16). AFC, by contrast, transforms the verification task into itemized fact-checking: clinicians confirm whether each claim is supported by the linked guideline passage. This shifts the cognitive burden from holistic reasoning assessment to discrete correspondence checking, leveraging clinicians' familiarity with clinical guidelines and their own knowledge.

In this randomized controlled trial, AFC was compared against traditional transparency mechanisms on clinician trust in AI-generated oncology recommendations spanning diagnostic classification and treatment decisions. Clinicians from three specialties directly involved in cancer care were enrolled, and trust was evaluated across diverse cancer types.

# Methods

### Study Design

This was a prospective, randomized, controlled trial comparing five transparency conditions for presenting AI-generated treatment recommendations in oncology. Participants were first randomized to one of two arms (with or without natural language explanations) and then sub-randomized within each arm to a specific presentation format. The arm without explanations included Groups 1 and 3; the arm with explanations included Groups 2, 4, and 5 (AFC). The trial was approved by the institutional review board (2024-590-S-CB) and followed CONSORT guidelines for reporting randomized trials.

### Setting and Participants

Participants were recruited online through professional medical networks, specialty society listservs, and targeted social media outreach between May and December 2025. Eligible participants were licensed healthcare professionals actively practicing in one of three specialty groups: diagnostic radiology, medical oncology, or radiation oncology. All participants were required to have clinical responsibilities involving cancer diagnosis, staging, or treatment planning. Exclusion criteria included participation in development of the evaluated AI system and prior exposure to the AFC methodology.

**Interventions**

Participants evaluated 21 clinical cases specific to their medical specialty, with all cases presented in randomized order within the survey. Cases spanned seven cancer types with three cases per type: prostate cancer, breast cancer, lung cancer, colorectal cancer, kidney cancer, liver cancer (including hepatocellular carcinoma), and lymphoma. Each case included standardized clinical information comprising patient demographics, relevant medical history, tumor characteristics, staging information, and associated imaging findings.

AI-generated recommendations were produced using GPT-4.5 (OpenAI) with specialty-specific prompting, few-shot examples, and validated by board-certified specialists who confirmed clinical appropriateness and guideline concordance. The five presentation groups presented these recommendations as follows:

- **Group 1 (recommendation only):** The AI-generated treatment recommendation was presented without additional context. Recommendations included the suggested treatment approach with relevant clinical parameters (e.g., “Recommendation: PI-RADS Category 5 - High suspicion for clinically significant cancer” for radiology; “Recommendation: Active surveillance” for oncology; “Recommend intensity-modulated radiation therapy with a dose of at least 70 Gy combined with 2-3 years of androgen deprivation therapy” for radiation oncology).
- **Group 2 (recommendation + explanation):** The recommendation was accompanied by a natural language explanation of the clinical reasoning. Explanations described key factors considered in the recommendation, including tumor characteristics, patient factors, and the logic connecting these elements to the suggested approach. Explanations averaged 30-70 words.
- **Group 3 (recommendation + source citation):** The recommendation was accompanied by citations to relevant clinical practice guidelines (NCCN, ESMO, ASTRO, EAU, ACR, AASLD) or classification systems (PI-RADS, BI-RADS, Lung-RADS, LI-RADS,

Bosniak, Lugano). Each case included one to three citations with the specific guideline version and year.

- **Group 4 (recommendation + explanation + source citation):** The recommendation included both the natural language explanation and source citations, representing a comprehensive traditional transparency approach combining clinical rationale with evidence attribution.
- **Group 5 (recommendation + explanation + source citation + AFC):** The recommendation incorporated AFC verification, which decomposed each recommendation into discrete factual claims and linked each claim to the corresponding passage in source clinical guidelines, highlighted in the source document (guidelines listed above). (See Figure 1 for an example of G4 and G5)

**Randomization and Blinding**

Group assignment was embedded within the survey platform (SurveyMonkey, San Mateo, CA, USA), which also implemented the randomization sequence, and was concealed from participants until completion of baseline assessments. After initial verification that eligibility criteria were met, participants were randomly assigned to one of two arms (with or without explanations) and further sub-randomized to one of five presentation groups using a computer-generated randomization sequence with approximately equal allocation across the five presentation groups. Randomization was stratified by medical specialty to ensure balanced representation across arms. Participants remained blinded to the existence of alternative presentation formats throughout the study. Investigators analyzing primary outcomes were blinded to group allocation.

**Case Materials**

Case content was developed separately for each specialty to reflect discipline-specific clinical decision-making. Radiology cases (n=21) focused on diagnostic classification using standardized reporting systems: PI-RADS for prostate MRI, BI-RADS for breast imaging, Lung-RADS for lung cancer screening, LI-RADS for hepatocellular carcinoma, Bosniak classification for renal masses, TNM staging for colorectal cancer, and Lugano classification for lymphoma. Medical oncology cases (n=21) addressed systemic therapy decisions including treatment selection, molecular testing indications, and management of metastatic disease based on guidelines from ASCO, ESMO,

NCCN, AUA, and EAU. Radiation oncology cases (n=21) focused on treatment planning decisions including technique selection, dose-fractionation schemes, target volume delineation, and integration with systemic therapy based on guidelines from ASTRO, ESMO, EAU, ASCO, AUA, ILROG, and specialty consensus statements (all cases are provided in the supplemental material).

All cases derived from actual clinical scenarios that were de-identified and modified to ensure patient privacy while preserving clinical realism. Cases represented varying levels of complexity, ranging from straightforward guideline-concordant decisions to scenarios requiring nuanced clinical judgment.

### Outcomes

**Primary Outcome:** Trust in the AI recommendation measured using a 5-point Likert scale administered after each case (1 = do not trust at all; 2 = mostly distrust; 3 = neither trust nor distrust; 4 = mostly trust; 5 = completely trust). This scale was adapted from validated instruments for measuring trust in automated systems.

**Secondary Outcomes:** (1) Proportion of responses expressing trust, defined as a score of 4 or higher; (2) Effect modification by medical specialty, years of clinical experience, cancer type, and prior experience with AI or LLMs; (3) Number needed to treat (NNT) to achieve increased trust ratings by application of AFC.

### Statistical Analysis

The primary analysis compared each intervention group against Group 1 (recommendation only) as the reference condition using linear mixed-effects models. Trust score served as the dependent variable, with group assignment as the fixed effect and participant and case as crossed random effects to account for the repeated-measures structure (multiple cases per participant, multiple participants per case). Effect sizes were calculated as standardized mean differences (Cohen's d), computed as the estimated group difference from the linear mixed-effects model divided by the residual standard deviation (i.e., the within-cluster standard deviation after accounting for participant and case random effects). This model-based approach yields a different denominator than the total (marginal) standard deviation reported in descriptive statistics and is recommended for repeated-measures designs where between-cluster variance is substantial.

For the proportion expressing trust, generalized estimating equations with a logit link function and an exchangeable correlation structure were used to account for clustering within participants; the exchangeable structure was chosen because cases were presented in randomized order within the survey, making any temporal correlation pattern unlikely. Relative risks and risk differences were calculated with corresponding 95% confidence intervals. NNT was computed as the inverse of the absolute risk difference.

Pairwise comparisons among all five presentation groups used Tukey adjustment for multiple comparisons. Prespecified subgroup analyses examined potential effect modification by medical specialty and cancer type through inclusion of interaction terms; additional subgroup analyses by clinical experience, LLM usage, AI familiarity, practice setting, and geographic region were exploratory and should be interpreted with caution given the number of comparisons. Heterogeneity in treatment effects across subgroups was assessed using the Cochran Q statistic and I-squared, applied to subgroup-specific effect estimates analogous to a meta-analytic framework.

Sensitivity analyses included: (1) exclusion of participants with extreme baseline response patterns (all responses at floor or ceiling); (2) stratification by self-reported AI familiarity; and (3) analysis restricted to complete responders. All analyses followed the intention-to-treat principle. Missing item-level data (<2% of responses) were addressed using multiple imputation with 20 imputed datasets; 20 participants (5.6%) had at least one missing case assessment. All reported denominators and proportions reflect imputed totals based on the intention-to-treat population. Sample size was determined a priori based on detecting a small-to-medium effect size (Cohen's $d = 0.35$) between groups with 80% power at a two-sided alpha of 0.05. With two arms, five presentation groups, 21 cases per participant, and an assumed intraclass correlation of 0.15, approximately 300 participants were required. A total of 380 participants were enrolled, anticipating 15-20% attrition, with 356 (93.7%) completing all assessments and included in the primary analysis.

Two-sided $P < 0.05$ was considered statistically significant. Analyses were conducted using Python 3.11 (scipy, statsmodels) and R 4.3 (lme4, emmeans). As a sensitivity analysis, ordinal cumulative link mixed models were also fitted to confirm robustness of results to the ordinal nature of the Likert scale; results were substantively unchanged.

# Results

## Study Population

Between May and December 2025, 380 participants enrolled and randomized. A total of 356 participants (93.7%) completed the study and were included in the primary analysis (Figure 2). The analytic cohort comprised 160 radiologists (44.9%), 111 medical oncologists (31.2%), and 85 radiation oncologists (23.9%), generating 7,476 individual trust ratings across 21 clinical cases per participant. Baseline characteristics were well-balanced across the two randomization arms with no statistically significant differences in age, sex, specialty distribution, years of experience, practice setting, or prior AI exposure (Table 1; all P > .15).

Clinical experience was diverse: 38 participants (10.7%) were residents or fellows, 89 (25.0%) had 1-5 years of independent practice, 104 (29.2%) had 5-10 years, 101 (28.4%) had 11-20 years, and 24 (6.7%) had more than 20 years of experience. AI familiarity ranged from participants who had never heard of clinical AI applications (17 participants, 4.8%) to those reporting regular clinical AI use (140 participants, 39.3%). Large language model usage patterns similarly varied, with 78 participants (21.9%) reporting no prior LLM use and 94 (26.4%) reporting daily use.

## Overall Effect of Atomic Fact-Checking

Mean trust scores differed substantially between the AFC condition and all control conditions (Figure 3a). Traditional transparency approaches (Groups 1-4) significantly improved trust, although still clustering near the neutral midpoint of the 5-point scale: Group 1 (recommendation only), 2.59 (SD 1.01; 95% CI, 2.54-2.64); Group 2 (recommendation plus explanation), 2.84 (SD 0.99; 95% CI, 2.79-2.90); Group 3 (recommendation plus source citation), 3.01 (SD 0.97; 95% CI, 2.97-3.05); Group 4 (recommendation plus explanation and source), 3.09 (SD 0.98; 95% CI, 3.04-3.14). The pooled control mean was 2.89 (SD 1.01; 95% CI, 2.86-2.91). In contrast, Group 5 (AFC) achieved a mean trust score of 3.80 (SD 0.82; 95% CI, 3.76-3.84), representing a shift from slight distrust or neutrality to strong trust. This score approached the scale maximum of 5 (“completely trust”), with 19.4% of AFC responses at the ceiling.

## Effect Size Estimation

The primary effect size comparing AFC against pooled controls was Cohen’s d = 0.94 (95% CI, 0.88-1.00; P < .001), classified as “large” by conventional benchmarks (d > 0.8). The unstandardized mean difference was 0.91 points (95% CI, 0.86-0.96) on the 5-point scale. In the prespecified linear mixed-effects model with crossed random effects for participant and case, the AFC coefficient was 0.88 (SE 0.05; $t = 17.6$; $P < .001$). The comparison of Group 5 versus Group 4, which isolates the AFC component from accompanying explanation and citation, yielded Cohen's $d = 0.78$ (95% CI, 0.67-0.89; $P < .001$), confirming that AFC rather than the accompanying transparency features drove the trust increase.

The intraclass correlation coefficient for participants was 0.18 (95% CI, 0.14-0.22), indicating that 18% of variance in trust ratings was attributable to stable between-participant differences. The ICC for cases was 0.06 (95% CI, 0.03-0.10), suggesting relatively consistent effects across clinical scenarios. The model explained 35.0% of total variance (marginal $R^2 = 0.12$; conditional $R^2 = 0.35$).

**Comparisons Among Control Conditions**

Pairwise comparisons among the four control groups revealed a dose-response gradient, with small-to-medium effect sizes ($d = 0.08$ to $0.50$; Supplementary Fig. 3a, Table 2): adding explanations, citations, or both progressively increased clinician trust. The largest pairwise difference was between Group 1 (recommendation only) and Group 4 (explanation plus citation; $d = 0.50$), suggesting cumulative benefit from combining transparency approaches. Adding citations significantly improved trust, while adding explanations did not lead to a significant difference in pairwise comparisons. All pairwise control effect sizes were well below the AFC effect ($d = 0.94$).

**Response Distribution Analysis**

Examination of response distributions revealed distinct patterns between conditions (Supplementary Fig. 2). In control groups, responses were approximately normally distributed around the neutral midpoint, with 11.8% of responses at the floor or ceiling combined. Modal responses were "neither trust nor distrust" (score 3) for Groups 1, 2, and 4, and "mostly trust" (score 4) for Group 3.

The AFC condition showed a markedly right-skewed distribution with pronounced ceiling effects: 47.1% of responses were "mostly trust" (score 4), 19.4% were "completely trust" (score 5), 28.3% were neutral (score 3), and 5.3% expressed some distrust (scores 1-2). This ceiling effect suggests the true effect of AFC may be underestimated by the observed effect size.

**Proportion Expressing Trust**

The proportion of responses expressing trust (score ≥ 4) differed dramatically between conditions (Figure 3b). Control groups showed consistently low trust rates with overlapping confidence intervals: Group 1, 18.1% (274/1,512; 95% CI, 16.3-20.1); Group 2, 24.9% (334/1,344; 95% CI, 22.6-27.2); Group 3, 30.7% (561/1,827; 95% CI, 28.6-32.9); Group 4, 33.6% (466/1,386; 95% CI, 31.2-36.2). The pooled control trust rate was 26.9% (1,635/6,069; 95% CI, 25.8-28.1).

With AFC, 66.5% (935/1,407; 95% CI, 63.9-68.9) of responses expressed trust. Using generalized estimating equations, the relative risk for expressing trust with AFC versus pooled controls was 2.47 (95% CI, 2.33-2.61; $P < .001$). The absolute risk increase was 39.5 percentage points (95% CI, 36.5-42.6), yielding a number needed to treat of 2.53 (95% CI, 2.37-2.72).

**Effect Modification by Medical Specialty**

AFC effects were consistent across the three medical specialties with no significant heterogeneity (Supplementary Fig. 3b) ($I^2 = 30\%$; Cochran's $Q = 2.86$; $P = .24$). Point estimates were largest among radiologists ($d = 1.03$; 95% CI, 0.94-1.11), followed by medical oncologists ($d = 1.01$; 95% CI, 0.92-1.11) and radiation oncologists ($d = 0.80$; 95% CI, 0.69-0.90) (Figure 4a). The interaction between intervention and specialty was not statistically significant ($P = .18$).

Trust conversion rates followed a similar pattern (Figure 4b). Among radiologists ($n = 160$), trust increased from 21.9% (95% CI, 20.2-23.6) in control conditions to 63.9% (95% CI, 60.3-67.4) with AFC (absolute increase: 42.0 percentage points). Among radiation oncologists ($n = 85$), trust increased from 45.0% (95% CI, 42.2-47.8) to 81.8% (95% CI, 78.3-85.0), representing a 36.8 percentage point increase. Among medical oncologists ($n = 111$), trust increased from 20.4% (95% CI, 18.3-22.6) to 58.1% (95% CI, 53.6-62.5), an increase of 37.7 percentage points. Notably, radiation oncologists showed the highest baseline trust (45.0%) and the highest AFC trust rate (81.8%).

## Effect Modification by Clinical Experience

Effect sizes varied significantly by years of clinical experience (interaction P = .003). Residents and fellows showed the largest effect, while experienced clinicians showed consistent but slightly smaller effects across all experience levels (Figure 5a).

Residents and fellows (n = 38) showed the largest effect: d = 1.41 (95% CI, 1.24-1.58), with trust increasing from 26.1% to 84.4%. Clinicians with 11-20 years of experience (n = 101) showed d = 0.77 (95% CI, 0.67-0.86), with trust increasing from 27.5% to 59.7%. Clinicians with 5-10 years of experience (n = 104) showed d = 1.01 (95% CI, 0.91-1.11), with trust increasing from 27.6% to 70.6%. Early-career clinicians with 1-5 years of experience (n = 89) showed d = 1.02 (95% CI, 0.91-1.13), with trust increasing from 27.1% to 70.2%. Senior clinicians with more than 20 years of experience (n = 24) showed d = 0.62 (95% CI, 0.43-0.82), with trust increasing from 22.6% to 44.8%.

## Effect Modification by LLM Usage Frequency

LLM usage frequency was associated with effect magnitude (interaction P < .001) (Figure 5b). Effects were large across all usage patterns: non-users showed d = 0.91 (95% CI, 0.80-1.03), weekly users d = 0.88 (95% CI, 0.79-0.97), daily users d = 1.04 (95% CI, 0.93-1.15), and monthly-or-less users d = 0.98 (95% CI, 0.84-1.12).

Non-users of LLMs showed the smallest but still substantial effect (d = 0.91; 95% CI, 0.80-1.03). The variation in effect magnitude across usage levels likely reflects differences in baseline trust and familiarity with AI-generated content rather than differential responsiveness to atomic verification.

## Effect Modification by AI Familiarity

Effects were large regardless of prior AI experience, though magnitudes varied (interaction P = .02) (Figure 5c). Clinicians reporting research-only AI use showed the largest effect (d = 1.05; 95% CI, 0.93-1.18), followed by those using AI clinically less than monthly (d = 0.91; 95% CI, 0.79-1.04). Clinicians who had never heard of clinical AI showed a non-significant effect (d = 0.26; 95% CI, 0.00-0.53; P = .06), though this subgroup comprised only 3 AFC participants, precluding reliable estimation. Those using AI regularly in clinical practice showed d = 1.01 (95% CI, 0.92-1.09).

The pattern suggests that clinicians with some AI awareness but limited clinical AI experience are particularly receptive to atomic verification, while those with regular clinical AI use may have higher baseline trust, attenuating the observable effect.

**Effect Modification by Cancer Type**

AFC improved trust across all seven cancer types, with effect sizes ranging from d = 0.80 to d = 1.10 (Supplementary Fig. 1a). All cancer types showed statistically significant improvement with AFC (all $P < .001$). Heterogeneity across cancer types was moderate ($I^2 = 38\%$; Q = 7.2; P = .30), suggesting some variability in effect magnitude but consistent direction of benefit.

**Case-Level Consistency**

All 21 clinical cases individually demonstrated positive AFC effects, with case-specific effect sizes ranging from d = 0.71 to d = 1.20 (Supplementary Fig. 4). All 21 confidence intervals excluded zero, demonstrating robustness of the AFC effect across diverse clinical scenarios, cancer types, and specialty domains. The median case-level effect size was d = 0.94 (IQR, 0.86-1.03). Within-case heterogeneity was relatively low, indicating consistent participant responses across diverse clinical scenarios.

## Discussion

AFC produced a large increase in clinician trust toward AI-generated oncology recommendations (Cohen's d = 0.94), substantially exceeding conventional thresholds for clinical significance. Traditional transparency mechanisms, including natural language explanations (d = 0.25) and source citations (d = 0.42), showed only small-to-medium improvements over baseline recommendations. The proportion of clinicians expressing trust increased from 26.9% with traditional approaches to 66.5% with AFC, yielding a number needed to treat of 2.53. Importantly, this NNT reflects survey-measured trust rather than clinical recommendation uptake; whether the observed trust advantage translates into changed clinical behavior remains an open translational question. Because Group 5 included all transparency components (explanation, source citation, and AFC), the effect size for AFC versus pooled controls (d = 0.94) reflects the combined addition. The comparison of Group 5 versus Group 4, which isolates the AFC component, yielded a similarly large effect (d = 0.78, P < .001), confirming that AFC rather than the accompanying transparency features drove the observed trust increase. These findings challenge prevailing assumptions about explainability as the primary pathway to clinical AI trust (17, 18).

The dose-response effect of traditional transparency mechanisms is noteworthy. While substantial research has focused on explainability as a trust-building strategy (11, 19), these results suggest that textual explanations and source citations provide incremental but limited benefit for high-stakes clinical decisions. While explanations and citations improved trust over bare recommendations (d = 0.25-0.50), the incremental gains were modest compared to AFC. This aligns with prior observations about the limitations of explainability approaches (20). The relatively small effect of traditional transparency compared to AFC suggests a qualitative difference between explaining AI reasoning and enabling direct verification of AI claims (21).

The success of AFC may stem from a difference in verification modality. Traditional explainability asks clinicians to evaluate AI reasoning quality, a cognitively demanding task requiring both technical AI understanding and domain expertise. In Kahneman's dual-process framework, this engages effortful System 2 deliberation (16). AFC, by contrast, shifts verification toward a simpler factual-matching task—checking whether a specific claim appears in a cited guideline—that leverages clinicians' existing interpretive skills and may operate closer to rapid System 1 pattern recognition.

Rather than assessing whether AI reasoning sounds plausible, clinicians can verify whether each claim is directly supported by the linked guideline passage. This reduction in cognitive load may explain why AFC produces substantially larger trust effects than explanation-based approaches across all subgroups examined.

This mechanism connects to the broader concept of augmented intelligence, which emphasizes human-AI collaboration over automation (22, 23). Studies of AI-assisted diagnosis in radiology, dermatology, and pathology have demonstrated that AI augmentation improves clinician performance when appropriately designed (24, 25). The current findings extend this paradigm by suggesting that verification-based trust mechanisms are fundamentally more effective than explanation-based approaches. Clinicians appear more willing to incorporate AI recommendations when they can confirm, rather than merely accept, the underlying basis.

Several clinical implications follow from these findings. First, investment in explainability features, while valuable for debugging and auditing (26), appears insufficient for achieving clinical adoption. Development priorities should shift toward verification interfaces that allow clinicians to confirm AI outputs against source guideline evidence. For digital medicine platforms deploying clinical decision support, this represents an actionable design principle: embed verifiability at the interface level rather than relying on post-hoc explanations. Second, AI systems should ground recommendations in information that clinicians can independently assess. For oncology, this means linking treatment recommendations to trusted medical guidelines or evidence.

The consistency of effects across specialties is notable. Radiologists, radiation oncologists, and medical oncologists all showed large effects (d = 0.80 to 1.03), indicating that AFC extends beyond imaging specialists. The slightly larger effect among radiologists (d = 1.03) may reflect their high trust towards sources represented in the AFC visualization, making visual verification particularly valuable for confirming recommendations outside their primary expertise. Radiation oncologists showed the highest baseline trust (45.0%) and the highest AFC trust rate (81.8%), possibly reflecting routine use of oncological guidelines and an extended knowledge about medical AI systems (27). The structured presentation of each claim alongside its source passage engages active verification rather than passive reception. The approach leverages existing guideline interpretation expertise that clinicians have already developed.

The AFC visualization used in this trial presented each atomic fact separately alongside the linked guideline passage. Alternative formats (e.g., in-line citations) may be effective, though direct linking to the source passage appears essential for enabling independent verification.

The present study has several limitations. Trust was measured via self-report in a survey context, which may systematically overestimate real-world adoption. The well-documented intention-behavior gap suggests that stated trust does not directly predict recommendation uptake in clinical workflows with competing time pressures, liability considerations, and patient preferences. Establishing ecological validity through studies measuring actual recommendation incorporation in clinical practice is a critical next step. The study was conducted across three subspecialties, and generalizability to other medical domains requires investigation. Immediate trust responses were assessed, but durability over repeated exposures was not evaluated. All AI recommendations were pre-validated as correct by expert panels, leaving trust calibration under error conditions untested. This is a critical gap: if AFC inflates trust even when the underlying recommendation is incorrect, the mechanism could promote overtrust rather than appropriate trust. Studying whether AFC preserves clinicians' ability to identify and reject erroneous recommendations is arguably more important for patient safety than demonstrating trust under correctness, and should be a priority for future work. The sample was predominantly from North American and European healthcare professionals, limiting generalizability to other healthcare settings. The 5-point Likert scale produced pronounced ceiling effects in the AFC group (19.4% at score 5), which compress the observed standard deviation and may inflate standardized effect sizes; the true magnitude of the AFC effect may therefore differ from what the bounded scale can capture. The null finding for clinicians who had never heard of AI in medicine ($d = 0.26$, $P = .06$, $n = 3$ in AFC arm) suggests that AFC's mechanism may presuppose some baseline awareness of AI capabilities; clinicians without this frame of reference may not engage with the verification process as intended, though the extremely small subgroup size precludes firm conclusions. While consistent effects across subgroups were observed, the mechanisms underlying variation in effect magnitude require further investigation, since performed Likert Scale-assessment does not provide insides regarding individual reasoning behind trust score ratings. Multiple subgroup interaction tests were performed, and findings from exploratory subgroup analyses (e.g., AI familiarity, interaction $P = .02$) should be interpreted cautiously, as they would not survive correction for multiple comparisons.

Verification-based trust mechanisms outperform explanation-based approaches by a wide margin in this clinical context. Whether this advantage extends to other medical domains, holds under conditions of varying AI accuracy, and translates into changed clinical behavior remain open questions. The immediate practical implication is that clinical AI systems should prioritize verifiability over explainability when the goal is appropriate clinician trust. As digital medicine platforms increasingly integrate large language models into clinical workflows, designing for verification rather than explanation may be the most effective path toward responsible AI adoption in healthcare.

**Author Contributions**

LA: Conceptualization, study design, formal analysis, statistical analysis, funding acquisition, and writing of the original draft. JCP: Conceptualization, study design, formal analysis, funding acquisition, and manuscript writing and revision. ETO: Data acquisition and critical revision of the manuscript. FM: Funding acquisition and critical revision of the manuscript. SEC and MRM: Provision of infrastructure and critical revision of the manuscript. LM, KB, SZ, DB, and MG: Critical revision and proofreading of the manuscript. All authors read and approved the final manuscript.

**Acknowledgements**

JCP and FM were funded by Google.org Gen AI Accelerator 2025. The funder played no role in study design, data collection, analysis and interpretation of data, or the writing of this manuscript.

**Competing Interests**

JCP holds shares in Mevidence GmbH and received honoraria from Brainlab and AstraZeneca. FM holds shares in Mevidence GmbH. KB reports grants from the German Federal Ministry of Research, Technology and Space, Free State of Bavaria (Bayern Innovativ), Wilhelm Sander Foundation, Max Kade Foundation and the European Union; speaker fees from Canon Medical Systems and GE Healthcare; and Data Safety Monitoring Board participation for Philips (IHI Project IMAGIO).
The remaining authors declare no financial or non-financial competing interests.

**Data Availability**

The case vignettes are shared in the supplement.

**Code Availability**
The underlying code for this study is not publicly available but may be made available to qualified researchers on reasonable request from the corresponding author.

**Ethics Approval**
This study was approved by the institutional review board of TUM University Hospital (approval number 2024-590-S-CB).

**Informed Consent**
All participants provided informed consent prior to enrollment.

## Figures

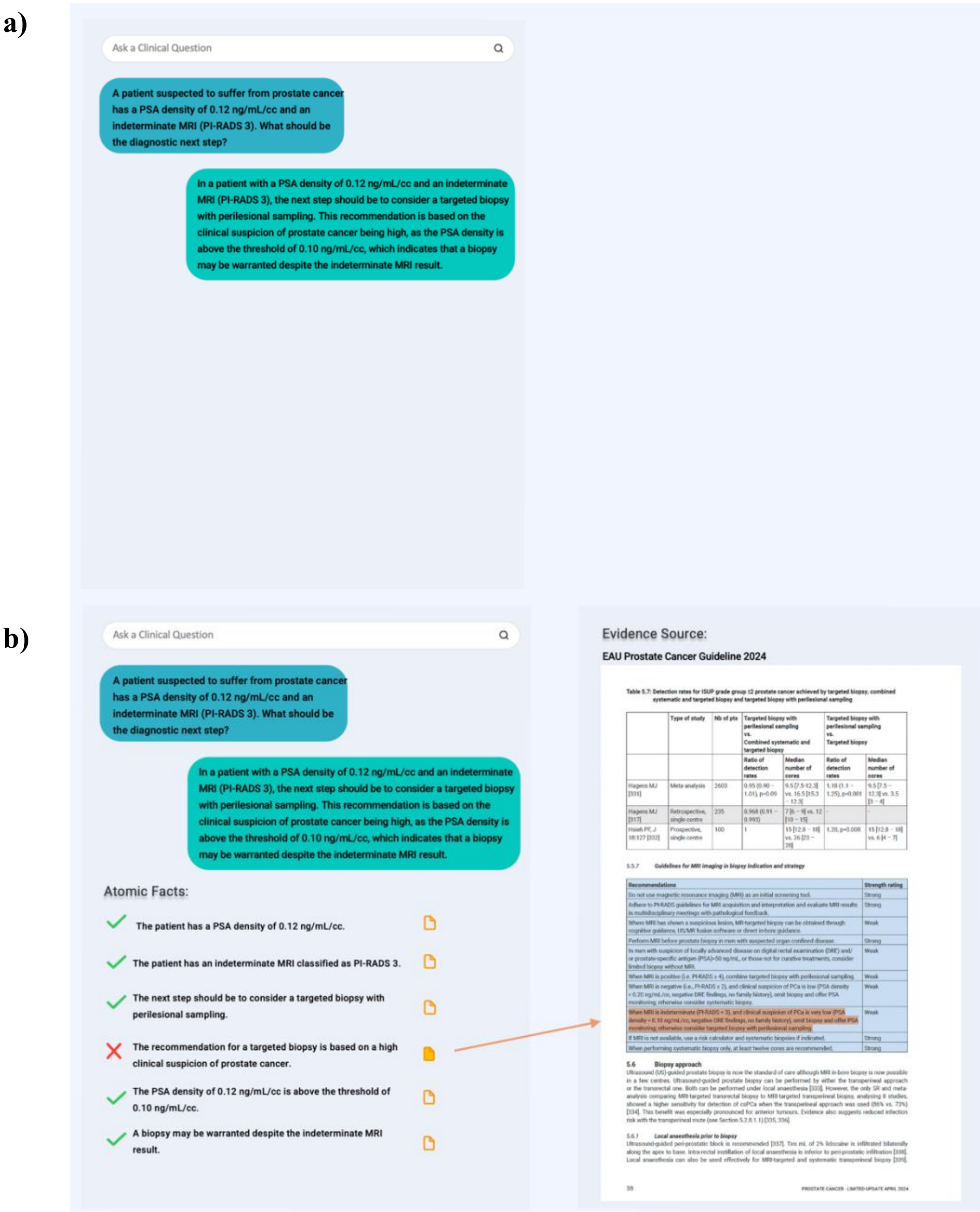

**Fig. 1 | Exemplary user interface of recommendations. a** Example of Group 2 (recommendation + explanation) presentation. **b** Example of Group 5 (recommendation + explanation + source citation + Atomic Fact-Checking) presentation.

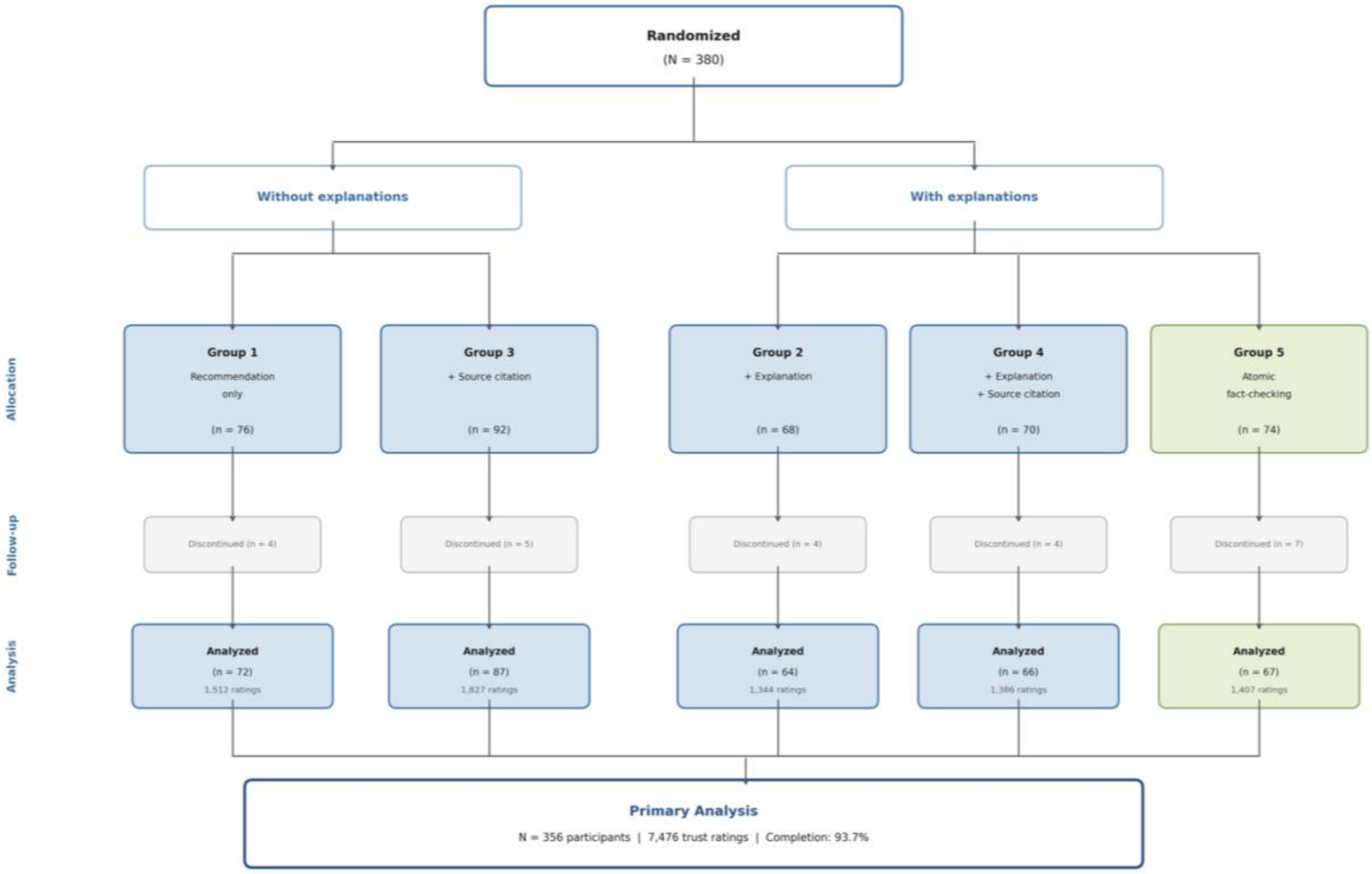

**Fig. 2 | CONSORT flow Diagram.** Of 380 enrolled participants, 356 (93.7%) completed the study and were included in the primary analysis. Participants were randomly assigned to one of two arms (with or without explanations) and sub-randomized to one of five presentation groups using a computer-generated randomization sequence with approximately equal allocation across the five presentation groups. Each participant evaluated 21 oncology cases, generating 7,476 total trust ratings.

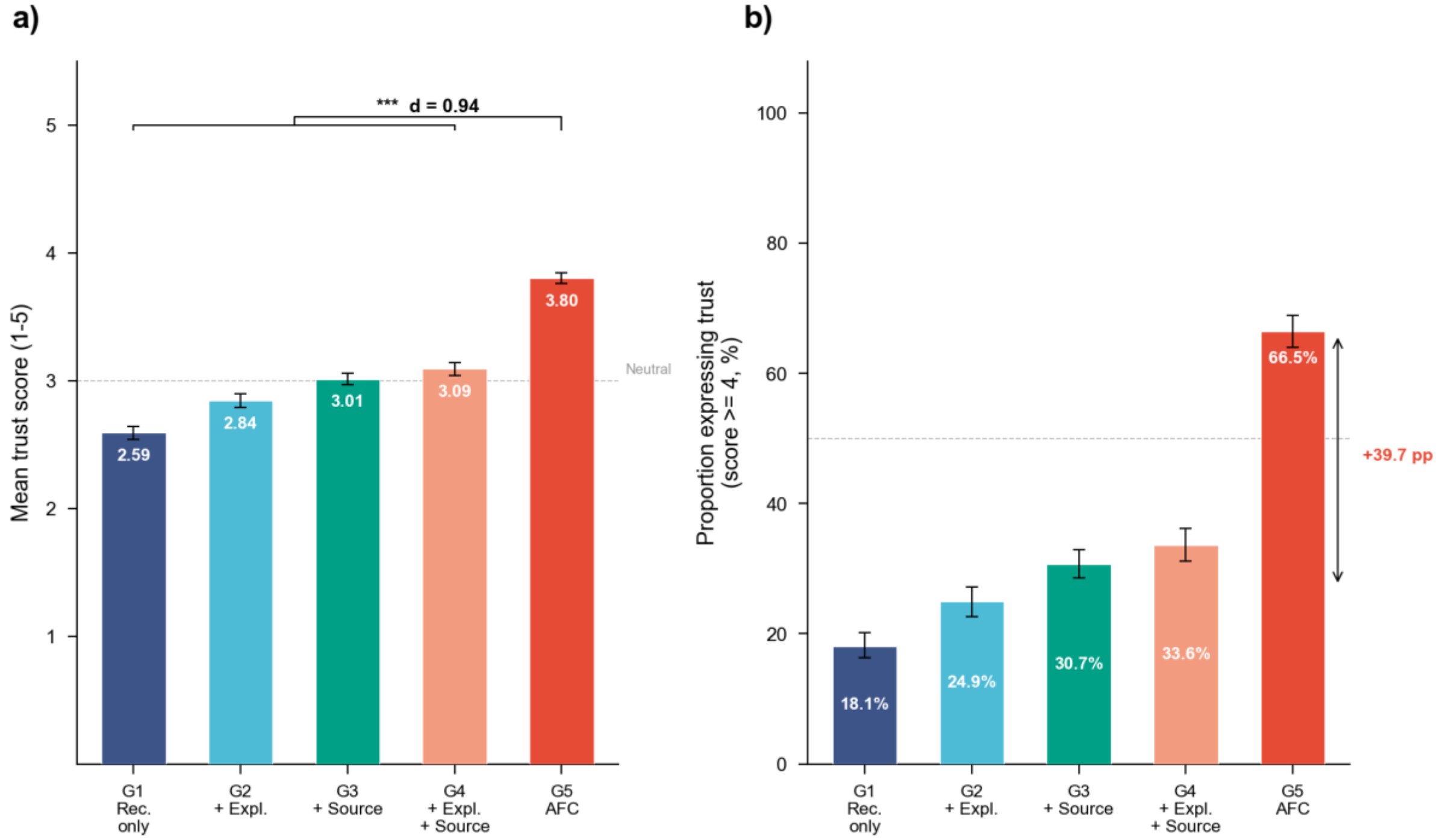

**Fig. 3 | Trust ratings. a** Primary outcome: mean trust score by experimental group. Bars show mean trust scores on the 5-point Likert scale (1 = do not trust; 5 = completely trust); error bars indicate 95% confidence intervals. A dashed line marks the neutral midpoint (3.0). Groups 1-4 (traditional transparency approaches) cluster near the neutral point, while Group 5 (atomic fact-checking) shows notably higher trust (3.80). Bracket indicates Cohen's d = 0.94 (95% CI, 0.88-1.00; ***P < .001) for Group 5 versus pooled Groups 1-4. **b** Proportion of clinicians expressing trust by experimental group. Trust defined as a rating ≥4 (mostly trust or completely trust). Bars show proportions with 95% Wilson score confidence intervals; percentages are displayed inside each bar. Control groups (G1-G4) showed trust rates of 18-34%, while AFC (G5) achieved 66.5%. Arrow indicates the absolute risk increase of 39.5 percentage points compared to pooled controls. A dashed line marks the 50% threshold. G1 = recommendation only; G2 = + explanation; G3 = + source citation; G4 = + explanation + source; G5 = atomic fact-checking.

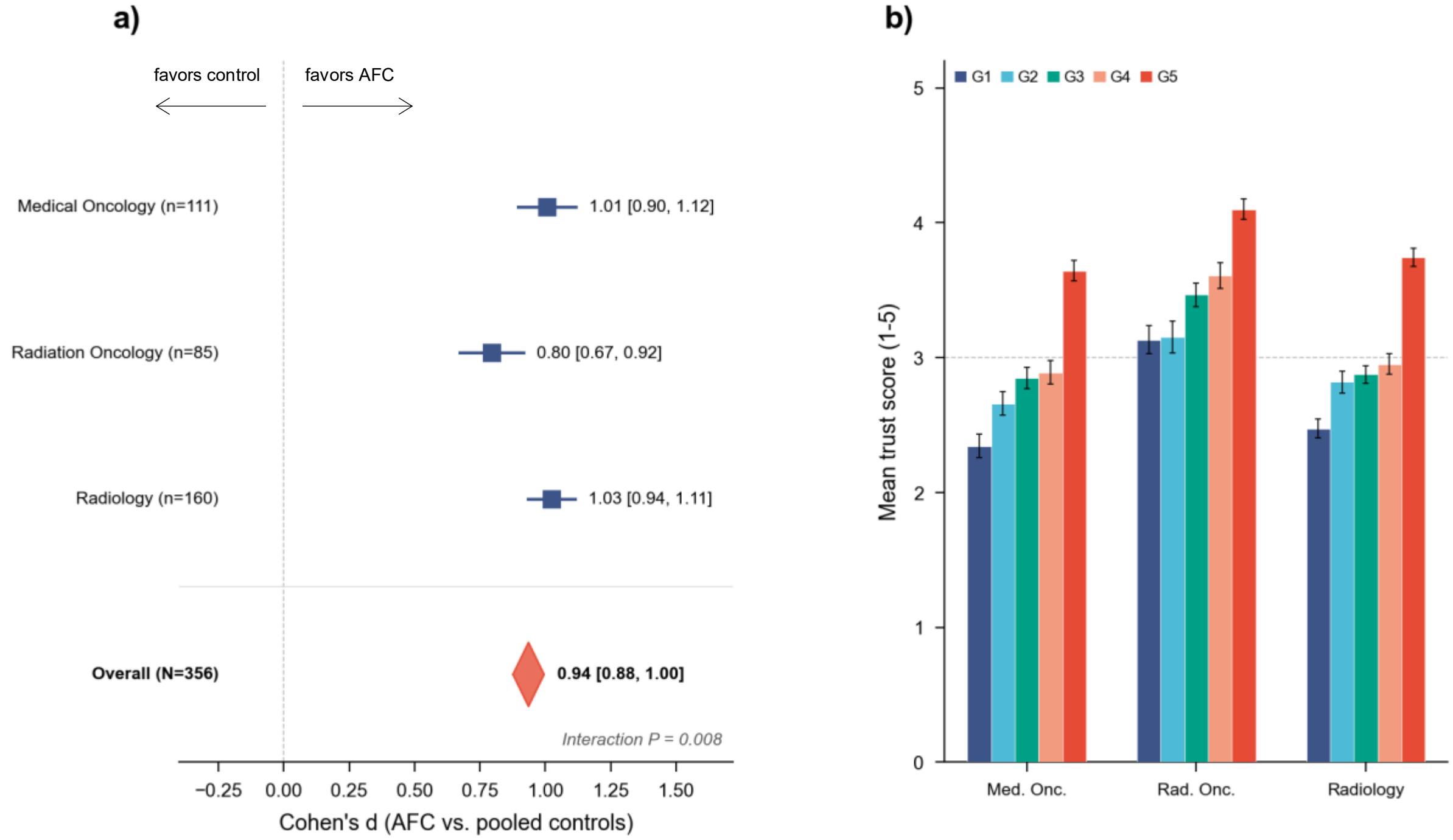

**Fig. 4 | Trust score effect modification by medical specialty. a** Forest plot: effect size by medical specialty. Squares indicate point estimates for Cohen's d comparing atomic fact-checking versus pooled controls; horizontal lines show 95% confidence intervals. The diamond indicates the pooled overall effect ($d = 0.94$). A dashed vertical line marks the overall estimate. All specialties showed large effects ($d \geq 0.80$) with no significant heterogeneity ($I^2 = 30\%$, $P = .24$). **b** Mean trust score by specialty and experimental group. Grouped bars show mean trust scores for each of five experimental groups within each specialty; error bars indicate 95% confidence intervals. A dashed line marks the neutral midpoint (3.0). The AFC effect was consistent across specialties ($d = 0.80$-1.03), with trust proportions increasing from 20-45% in control groups to 58-82% with AFC.

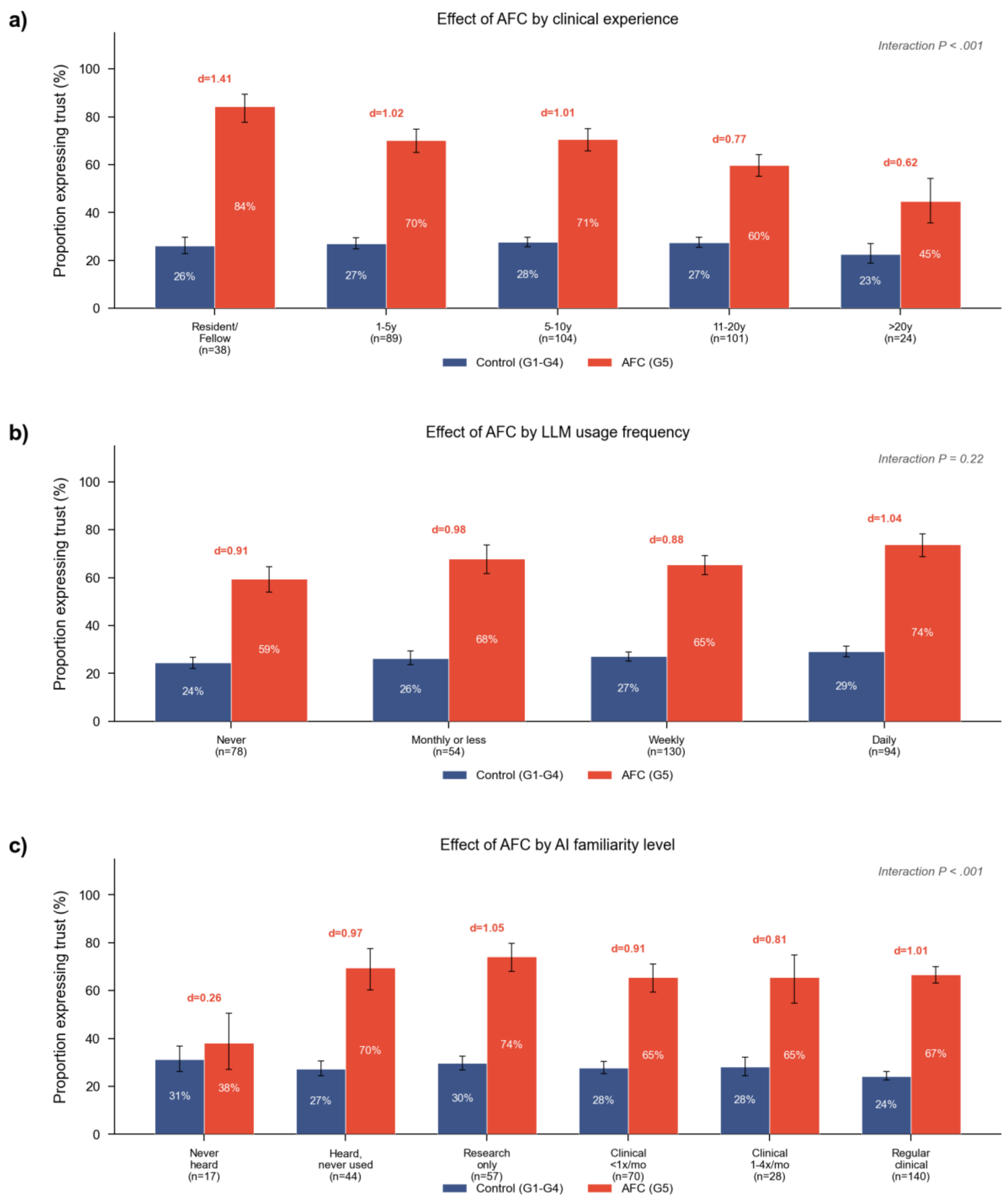

**Fig. 5 | Effect of atomic fact-checking versus control. a** Effect of atomic fact-checking by clinical experience. Grouped bars compare the proportion expressing trust (≥4) between control (dark blue) and AFC (red) conditions across five experience levels. Error bars indicate 95% Wilson score confidence intervals. Percentages are displayed inside bars; Cohen's d values are shown above each AFC bar. Sample sizes (n) are shown below each category. Largest effect in

residents/fellows (d = 1.41); consistent effects across all other experience levels (d = 0.62-1.41). Interaction P = .003. **b** Effect of atomic fact-checking by LLM usage frequency. Grouped bars compare trust proportions across four usage levels. Effects were consistent across all usage patterns (d = 0.88-1.04). Interaction P < .001. **c** Effect of atomic fact-checking by AI familiarity level. Grouped bars compare trust across six levels of prior AI experience. AFC improved trust across most familiarity levels (d = 0.26-1.05), though the effect was non-significant for AI-naive clinicians (n = 3 clinicians), with the largest effects among clinicians with research-only AI experience (d = 1.05) and those reporting regular clinical AI use (d = 1.01). Interaction P = .02.

## Tables

### Table 1 | Baseline characteristics of study participants

| Characteristic | Group 1 (n=72) | Group 2 (n=64) | Group 3 (n=87) | Group 4 (n=66) | Group 5 (n=67) | Total (N=356) |
|---|---|---|---|---|---|---|
| Age, median (IQR), y | 38 (32-43) | 36 (31-43) | 40 (34-47) | 39 (33-46) | 40 (33-52) | 38 (33-47) |
| **Sex** | | | | | | |
| Female | 30 (41.7) | 21 (32.8) | 40 (46.0) | 23 (34.8) | 28 (41.8) | 142 (39.9) |
| Male | 42 (58.3) | 43 (67.2) | 47 (54.0) | 43 (65.2) | 39 (58.2) | 214 (60.1) |
| **Medical specialty** | | | | | | |
| Radiology | 34 (47.2) | 27 (42.2) | 40 (46.0) | 28 (42.4) | 31 (46.3) | 160 (44.9) |
| Medical oncology | 21 (29.2) | 22 (34.4) | 26 (29.9) | 22 (33.3) | 20 (29.9) | 111 (31.2) |
| Radiation oncology | 17 (23.6) | 15 (23.4) | 21 (24.1) | 16 (24.2) | 16 (23.9) | 85 (23.9) |
| **Practice setting** | | | | | | |
| Academic medical center | 42 (58.3) | 39 (60.9) | 46 (52.9) | 33 (50.0) | 38 (56.7) | 198 (55.6) |
| Community practice | 30 (41.7) | 25 (39.1) | 41 (47.1) | 33 (50.0) | 29 (43.3) | 158 (44.4) |
| **Clinical experience** | | | | | | |
| Resident/fellow | 6 (8.3) | 8 (12.5) | 12 (13.8) | 4 (6.1) | 8 (11.9) | 38 (10.7) |
| 1-5 y | 21 (29.2) | 21 (32.8) | 17 (19.5) | 16 (24.2) | 14 (20.9) | 89 (25.0) |
| 5-10 y | 24 (33.3) | 16 (25.0) | 24 (27.6) | 24 (36.4) | 16 (23.9) | 104 (29.2) |
| 11-20 y | 17 (23.6) | 19 (29.7) | 24 (27.6) | 19 (28.8) | 22 (32.8) | 101 (28.4) |
| >20 y | 4 (5.6) | 0 (0.0) | 10 (11.5) | 3 (4.5) | 7 (10.4) | 24 (6.7) |
| **AI familiarity** | | | | | | |
| Never heard of clinical AI | 4 (5.6) | 1 (1.6) | 7 (8.0) | 2 (3.0) | 3 (4.5) | 17 (4.8) |
| Heard of but never used | 8 (11.1) | 13 (20.3) | 11 (12.6) | 4 (6.1) | 8 (11.9) | 44 (12.4) |
| Used in research only | 9 (12.5) | 6 (9.4) | 15 (17.2) | 19 (28.8) | 8 (11.9) | 57 (16.0) |
| Clinical use <1x/mo | 15 (20.8) | 19 (29.7) | 16 (18.4) | 8 (12.1) | 12 (17.9) | 70 (19.7) |
| Clinical use 1-4x/mo | 6 (8.3) | 4 (6.2) | 6 (6.9) | 6 (9.1) | 6 (9.0) | 28 (7.9) |
| Regular clinical use | 30 (41.7) | 21 (32.8) | 32 (36.8) | 27 (40.9) | 30 (44.8) | 140 (39.3) |
| **LLM usage frequency** | | | | | | |
| Never | 16 (22.2) | 10 (15.6) | 17 (19.5) | 18 (27.3) | 17 (25.4) | 78 (21.9) |
| Monthly or less | 12 (16.7) | 12 (18.8) | 17 (19.5) | 5 (7.6) | 8 (11.9) | 54 (15.2) |
| Weekly | 24 (33.3) | 21 (32.8) | 33 (37.9) | 26 (39.4) | 26 (38.8) | 130 (36.5) |
| Daily | 20 (27.8) | 21 (32.8) | 20 (23.0) | 17 (25.8) | 16 (23.9) | 94 (26.4) |

Data are n (%) unless otherwise indicated. Characteristics were balanced across randomization arms (all standardized differences <0.1). AI indicates artificial intelligence; IQR, interquartile range; LLM, large language model.

## Table 2 | Primary and secondary outcomes

### Panel A: Trust Scores and Proportions by Experimental Group

| Group | n | Trust Score, Mean (SD) | Trust Score, 95% CI | Proportion Expressing Trust, % (95% CI) |
|---|---|---|---|---|
| Group 1 (Recommendation only) | 72 | 2.59 (1.01) | 2.54-2.64 | 18.1 (16.3-20.1) |
| Group 2 (+ Explanation) | 64 | 2.84 (0.99) | 2.79-2.90 | 24.9 (22.6-27.2) |
| Group 3 (+ Source citation) | 87 | 3.01 (0.97) | 2.97-3.05 | 30.7 (28.6-32.9) |
| Group 4 (+ Explanation + Source) | 66 | 3.09 (0.98) | 3.04-3.14 | 33.6 (31.2-36.2) |
| *Pooled controls (Groups 1-4)* | 289 | 2.89 (1.01) | 2.86-2.91 | 26.9 (25.8-28.1) |
| **Group 5 (Atomic fact-checking)** | 67 | 3.80 (0.82) | 3.76-3.84 | 66.5 (63.9-68.9) |

### Panel B: Effect Sizes for Atomic Fact-Checking Versus Control Conditions

| Comparison | Cohen's d (95% CI) | Mean Difference (95% CI) | P Value | NNT (95% CI) |
|---|---|---|---|---|
| **AFC vs Pooled controls** | 0.94 (0.88-1.00) | 0.91 (0.86-0.96) | <.001 | 2.53 (2.37-2.72) |
| AFC vs Group 1 | 1.31 (1.23-1.39) | 1.21 (1.14-1.27) | <.001 | 2.07 (1.94-2.21) |
| AFC vs Group 2 | 1.05 (0.97-1.13) | 0.96 (0.89-1.03) | <.001 | 2.40 (2.22-2.62) |
| AFC vs Group 3 | 0.87 (0.79-0.94) | 0.79 (0.73-0.85) | <.001 | 2.80 (2.56-3.08) |
| AFC vs Group 4 | 0.78 (0.70-0.86) | 0.71 (0.64-0.78) | <.001 | 3.05 (2.75-3.41) |

### Panel C: Pairwise Comparisons Among Control Groups

| Comparison | Cohen's d (95% CI) | Mean Difference (95% CI) | P Value |
|---|---|---|---|
| Group 1 vs Group 2 | −0.25 (−0.32 to −0.18) | −0.25 (−0.32 to −0.18) | .28 |
| Group 1 vs Group 3 | −0.42 (−0.49 to −0.35) | −0.42 (−0.49 to −0.35) | <.001 |
| Group 1 vs Group 4 | −0.50 (−0.57 to −0.43) | −0.50 (−0.57 to −0.43) | .008 |
| Group 2 vs Group 3 | −0.17 (−0.24 to −0.10) | −0.17 (−0.24 to −0.10) | <.001 |
| Group 2 vs Group 4 | −0.25 (−0.33 to −0.18) | −0.25 (−0.32 to −0.17) | <.001 |
| Group 3 vs Group 4 | −0.08 (−0.15 to −0.01) | −0.08 (−0.15 to −0.01) | .16 |

Trust score measured on 5-point Likert scale (1 = do not trust at all; 5 = completely trust). Proportion expressing trust defined as score ≥4. Effect sizes classified as small (d = 0.2), medium (d = 0.5), large (d ≥ 0.8). P values are from linear mixed-effects models with participant and case as crossed random effects; pairwise comparisons adjusted using Tukey method. AFC indicates atomic fact-checking; CI, confidence interval; NNT, number needed to treat.